\documentclass[runningheads]{llncs}
\usepackage{graphicx}
\usepackage{wrapfig}
\usepackage{tikz}
\usepackage{comment}
\usepackage{amsmath,amssymb} 
\usepackage{color}
\usepackage{caption}
\usepackage{graphicx}
\usepackage{float} 
\usepackage{subcaption}
\usepackage{algorithm}
\usepackage{algpseudocode}
\usepackage[misc]{ifsym}
\usepackage{subcaption}

\usepackage[accsupp]{axessibility}  

\begin{document}
\pagestyle{headings}
\mainmatter
\def\ECCVSubNumber{7}  

\title{
Detect and Approach: Close-Range Navigation Support for People with Blindness and Low Vision
} 

\titlerunning{Detect and Approach}

\author{ Yu Hao\inst{1,2}\and
Junchi Feng\inst{2} \and
John-Ross Rizzo\inst{2,4} \and
Yao Wang\inst{2} \and
Yi Fang\inst{1,2,3^{(\textrm{\Letter})}}}

\authorrunning{Y. Hao et al.}

\institute{NYU Multimedia and Visual Computing Lab\\
NYU Tandon School of Engineering, New York University, USA 
\\
New York University Abu Dhabi, UAE\\
NYU Langone Health, USA\\
\email{\{yh3252, jf4151, yw523, yfang\}@nyu.edu}\\
\email{\{JohnRoss.Rizzo\}@nyulangone.org}}

\maketitle

\begin{abstract}
People with blindness and low vision (pBLV) experience significant challenges when locating final destinations or targeting specific objects in unfamiliar environments. 
Furthermore, besides initially locating and orienting oneself to a target object, approaching the final target from one's present position  is often frustrating and challenging, especially when one drifts away from the initial planned path to avoid obstacles. In this paper, we develop a novel  wearable navigation solution to provide real-time guidance for a user to approach a target object of interest efficiently and effectively in unfamiliar environments.  Our system contains two key visual computing  functions: initial target object localization in 3D and continuous estimation of the user's trajectory, both based on the  2D video captured by a low-cost monocular camera mounted on in front of the chest  of the user.  These functions enable the system to suggest an initial navigation path, continuously update the path as the user moves, and offer timely recommendation about the correction of the user's path. Our experiments demonstrate that our system is able to operate with an error of less than 0.5 meter both outdoor and indoor. The system is entirely vision-based and does not need other sensors for navigation, and the computation can be run  with the Jetson processor in the wearable system to facilitate real-time navigation assistance.


\end{abstract}
 
\section{Introduction}
According to 2020 WHO estimates, 295 million people suffer from moderate to severe visual impairment, while 43.3 million people are presently blind \cite{pascolini2012global}. Globally, between 1990 to 2020, the number of moderate to severely visually impaired increased by 91.7\%, and the number of people who were blind increased by 50.6\% \cite{hakobyan2013mobile}. This trend is predicted to continue with estimates approaching 474 million people with moderate to severe visual impairment and 61 million people with blindness by 2050 \cite{world2014visual}. Blindness and low vision poses significant challenges for nearly every activities of daily living  \cite{massiceti2018stereosonic}. One critical task element of most activities in daily living is visual search or a goal-oriented activity that involves the active scanning of the environment to locate a particular target among irrelevant distractors \cite{treisman1980feature}. Performing visual search can be demanding in complex environments, even for those with normal vision. It is even more challenging for the pBLV \cite{mackeben2011target}. For people with moderate to severe peripheral vision loss, central vision loss, and hemi-field vision loss, due to reductions  in the field of view, most have difficulty in isolating a particular location when searching for an object of interest and may need help in locating the object. For people experiencing  blurred vision or nearsightness, they may have difficulty in identifying object at relatively far distances. For people with color deficient vision and low contrast vision, it may be difficult for them to distinguish objects from background when the object and background share similar colors. Aside from isolating the particular location of an object, closing the distance between one's current position and the object itself is also a challenge. pBLV often want more than just information about the initial location of the object relative to their current position, but also continuous help in navigating to the  object along the way \cite{fernandes2019review}.

\begin{figure}[t] 
    \centering
    \includegraphics[width=11cm]{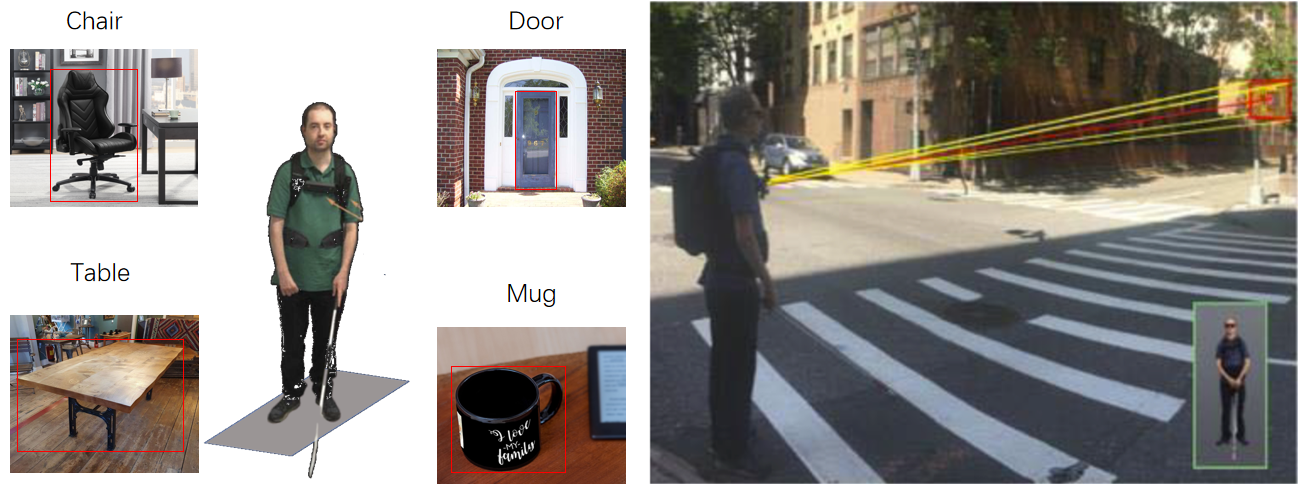}
    \caption{Challenges for pBLV exist in various scenarios: Searching for objects of interests and walking to  the objects (Left). Our wearable system contains a backpack with a monocular camera, an Nvidia Jetson Xavier NX Developer kit, and battery. The camera is placed on the  chest of the user (Right). }
    \label{fig2}
\end{figure}

\begin{figure}[t] 
    \centering
    \includegraphics[width=11cm]{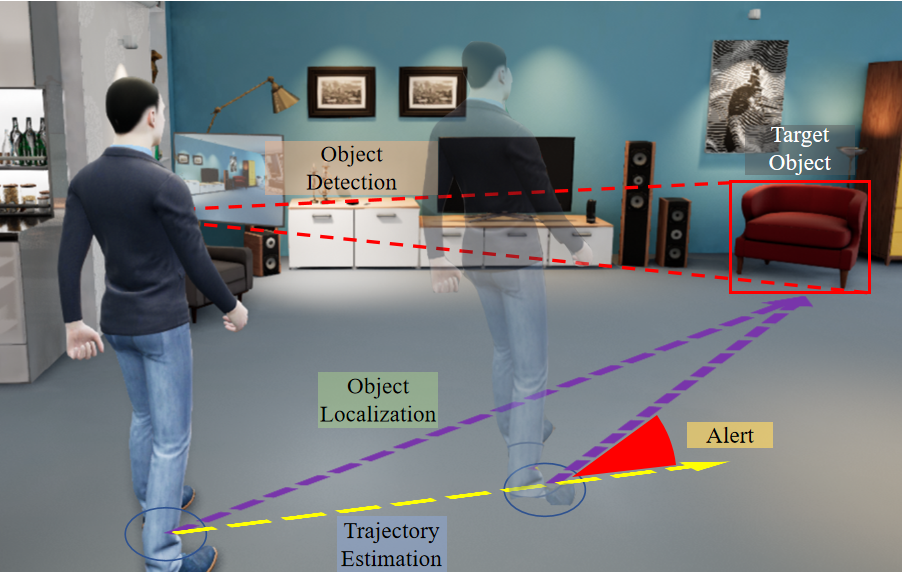}
    \caption{Our system is able to detect and locate an object of interest and guide a  person with BLV to  the target object. Initially, an object detection module will detect all possible interesting objects. Once the person selects a target object of interest, the object localization module will provide the 3D location of the object  and plan the path for the person to reach the object. The trajectory estimation module will then continuously estimate the person's movement between two time points, update the object location (relative to the user), and send path correction feedback to the user when necessary. 
    }
    \label{fig1}
\end{figure}

Solutions to aid this enormous and ever-growing problem of blindness and low vision are desperately needed. In the context of navigation and overcoming the close-range challenge, assistive technologies may help close the gap, and aid pBLV attain functional independence with better quality of life \cite{montello2009cognitive}. However, for the pBLV, only a limited number of tools have modest market traction and very few, if any, are able to support precise interaction with objects of interest in the surrounding environment.


Many of the present mobile apps for way-finding have decent success at leading end users to a general vicinity of a target location but few are able to precisely provide instructions as one approaches the target. As most apps  are focused on outdoor use and are predicated on GPS technology, they lack the accuracy required to support close-range navigation, which is necessary for pBLV to approach their final destination. Adding insult to injury, as most pBLV live in metropolitan environments, the accuracy of GPS-enabled smartphones reduces from a 4.9 meters radius under the open sky to a 20 meters radius in an urban setting, which is insufficient for reaching exact location and/or specific objects of interest \cite{foley2012technology}. 

In this work, we develop a new  wearable navigation solution to augment perceptive ability for pBLV. The system will  help a user to locate a target object of interest and provide guidance to reach the target efficiently and effectively in unfamiliar environments.
Our wearable system, as shown in Figure \ref{fig2}, contains a backpack with a monocular camera, an Nvidia Jetson Xavier NX Developer kit, and a battery. The camera is placed in front of the chest of the user in a custom scaffold that can be mounted on the shoulder strap of a backpack housing the Jetson board and the battery. With the sequence of images captured by the camera, our system is be able to detect the target object and provide real-time path planing and updated guidance to the end user as the user approaches the target object, with an accuracy of less than 0.5 meter.

The visual processing part of our system contains three main modules, as illustrated in Figure \ref{fig1}:   object detection, object localization, and trajectory estimation. The  object detection module is implemented by a pretrained YOLOv5 detection model \cite{YOLO}, which is responsible to detect all possible objects of interests. After the user selects an object as the target, the object localization module will provide the initial 3D coordinate of the target object and suggest an initial path for navigation from the user's current location to the target (e.g. the first purple path in the figure).  The  trajectory estimation module will then continuously  estimate the movement of the user (or more precisely the camera) and consequently update the desired path to the target (the second purple path in the figure).   If the angle between the updated  path and the estimated user's path (the yellow path) is higher than a pre-defined threshold, our system will send an alert message to the user. In this example, the system may say ``Please  head towards your left slightly by about 30 degrees".

\begin{figure}[t] 
    \centering
    \includegraphics[width=11cm]{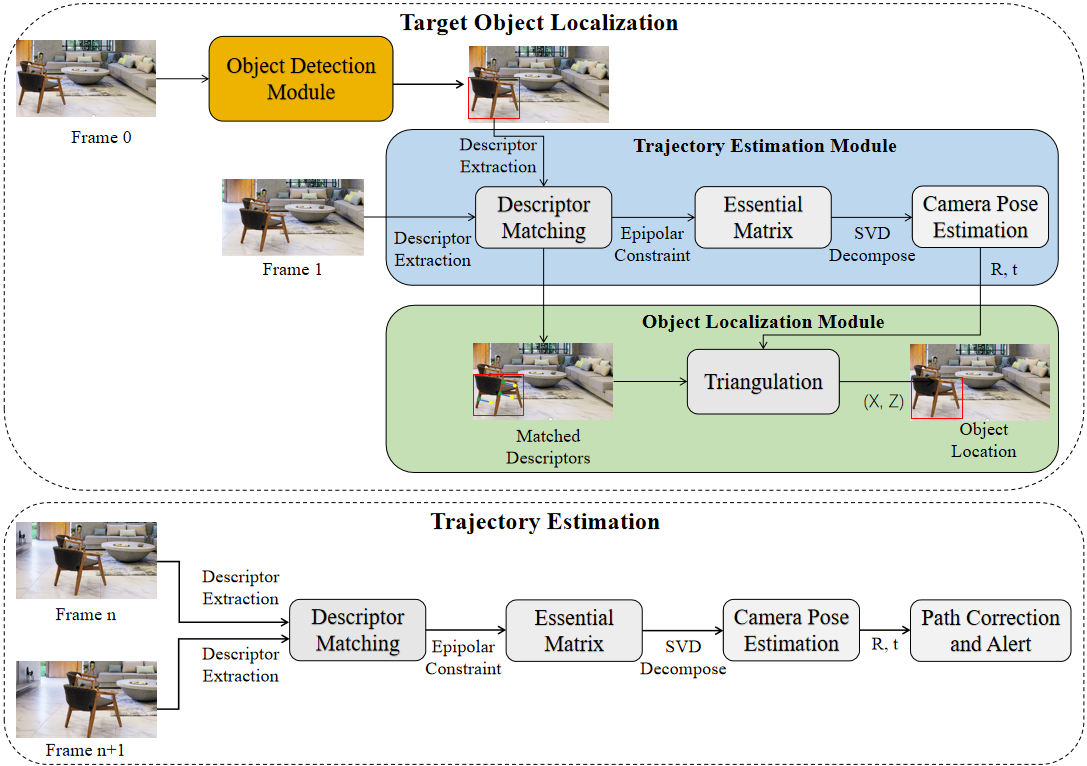}
    \caption{The processing flow for the  target object localization module (top) and trajectory estimation module (bottom). 
    }
    \label{fig3}
\end{figure}

To reduce the system cost and computational load,  we only use a deep-learning model for object detection in the first frame. We  estimate the initial 3D coordinates of the target object using the corresponding 2D locations of the object in two initial frames  as shown in Figure \ref{fig3}, to alleviate the need for a stereo camera for depth sensing. Given the initial position of the target object, we estimate the camera motion  between  successive frames to determine the  trajectory of the user, and update the object position relative to the user for continuous path updating as shown in Figure \ref{fig3}. 

Our experiments demonstrate that our system is able to detect objects of interests and provide real-time update of the object location relative to the  user as the user moves towards  the object, with an error of less than 0.5 meter. The system is entirely vision-based and does not need other sensors for navigation (e.g. IMUs and range sensors), and the computation can be run  with the Jetson processor in the wearable system to facilitate real-time navigation assistance.

      
  

\section{Related Works} 
Considering the growing prevalence of smartphones  in general and in  pBLV population \cite{liu2021virtual} \cite{lu2021audi} \cite{griffin2017survey}, mobile applications can be a potential solution to address the needs of localization and navigation for the pBLV \cite{real2019navigation}.  The All Aboard \cite{jiang2021field}, developed by Massachusetts Eye and Ear, utilizes computer vision to detect bus stop sign in the vicinity of the users and guide the users to the precise location of the bus stop by providing the distance estimations of the bus stop sign using  computer vision algorithms. The drawback of this app is  that it only detect the bus stop sign. 




Another example of the computer vision-based app for the pBLV is Virtual Touch \cite{liu2021virtual}. This app utilizes the smartphone's camera to capture the surrounding environment of the user, and detects objects of interest in the scene. This app enables the users to interact with the environment by pointing their fingers to an object of interest and the app will tell the users what object they are pointing to. However, this app  lacks the abilities of distance estimation and navigation.

Another category of assistive technology for navigation is light-based indoor positioning \cite{8950421}. This technology requires illuminating equipment such as LED light to illuminate the environment and transmit infrared signals at the same time. The user holds a receiver such as the smartphone to receive and decode the light signals. By calculating the angles of the received signals, it is possible to accurately localize the user and provide navigation for indoor locations where GPS doesn't work well. However, the cost of this system can be a concern as it requires significant efforts in establishing the qualified illuminating infrastructure.   

On the other hand, our system combines object detection and trajectory estimation features to provide all the functions necessary for detecting and navigating to a target object using a monocular camera only. It can detect different types of target, whereas the All Aboard app can only detect a specific type of object. Our system augments the pBLV's perceptive ability more than the  Virtual Touch app because our system also provides real-time path correction function to guide users to the target object. Moreover, our system is wearable and all the processing can be done locally,  and hence can provide navigation without the need of new infrastructure or other sensors except a monocular camera.

Our methodology for object localization and trajectory estimation from 2D video is similar to that used for visual SLAM  (simultaneous localization and mapping) \cite{qin2018vins} \cite{mur2015orb}, typically used to track the camera pose of a field robot and map the sounding environment relative to the robot. Here we use visual SLAM to estimate the movement of a user wearing a camera and the location of a particular stationary  object relative to the user. Therefore, the proposed system is an innovative integration of object detection and visual SLAM for assisting pBLV in detecting and approaching a target object.

\section{Methods}
As shown in Figure \ref{fig1},  our system consists of three main visual processing modules: object detection, 
object localization and trajectory estimation. We use a pretrained YOLOv5 model to detect all possible interesting objects on the first frame. After a user selects an object as the target object, the object localization module will determine the 3D coordinate of the object relative to the user (more precisely the camera center) based on first two frames of the video. This module is applied only in the first two frames  at the start of the navigation. Then the trajectory estimation module will continuously estimate the movement of the user between frames and consequently update the location of the target relative to the user's current position. Based on the updated object location, the system may provide path correction suggestions to the user. 
There are two options for selecting target object from all detected objects by the object detection module: 1) Use audio play all detected objects to the user and user uses the microphone to select the target object by existing audio to text API. 2) Use Virtual Touch \cite{liu2021virtual} to interact with the environment by pointing their fingers to the target object. The details of the object localization and trajectory estimation  modules are described in the following subsections. Because the object localization module makes use of the feature correspondence and camera motion estimation approaches used for trajectory estimation, we will first describe the trajectory estimation module in Sec.~\ref{sec2}, and then present the object localization module in Sec.~\ref{sec3}. 

\subsection{Trajectory Estimation} \label{sec2}
In this section, we introduce the trajectory estimation module, which aims to determine the movement of the user between two  video frames with a chosen frame interval.\footnote{The video is typically captured between 15 to 30 frames per second. But this processing may be done at a slower speed, e.g. every 0.5 to 1 second.} We make use of the fact that the camera is mounted in front of the user's chest and therefore the camera's movement is a good proxy for the user's movement. We adopt a classical approach for determining the rotation and translation of the camera between two camera views based on the correspondence of selected features points. The user's movement between the two frames is assumed to be equal to the estimated camera translation.  

Our trajectory estimation  module includes three  components: feature point detection and feature descriptor extraction; feature point  matching; and camera motion estimation based on the epipolar constraint, as shown in Figure \ref{fig3} (bottom). The following subsections describe these components.

\subsubsection{Feature Descriptor Extraction and Matching}

In recent years, many local feature detectors and descriptors, such as SIFT \cite{lowe2004distinctive},
SURF \cite{bay2008speeded} and ORB\cite{rublee2011orb}, have been developed and used for object recognition, image registration, classification, or 3D reconstruction. To enable real-time navigation  assistance, we chose the ORB feature descriptor \cite{rublee2011orb}, which are oriented multi-scale FAST \cite{alcantarilla2011fast} corners with a 256-bits descriptor associated. There are two main advantages of ORB: 1) ORB uses an orientation compensation mechanism, making it rotation invariant; 2) ORB learns the optimal sampling pairs, whereas other descriptors like BRIEF \cite{calonder2010brief} uses randomly chosen sampling pairs. These strategies boost the accuracy and efficiency of feature detection and matching. 
 
Based on the feature descriptors, we  establish the correspondences between the features in the current frame and the reference frame of the same scene. We first use a brute force matching algorithm to calculate the similarity between all descriptors in the current frame and all descriptors in the reference frame and determine an initial set of pairs of 2D coordinates of corresponding features. RANdom SAmple Consensus (RANSAC) \cite{fischler1981random} algorithm is then utilized to exclude the matching  outliers and furthermore estimate the essential matrix that best describes the geometric relation between corresponding 2D coordinates, to be  introduced in the next subsection.

\subsubsection{Determining the camera rotation and translation }
When a monocular camera views a 3D scene from two distinct positions and orientations, there are a number of geometric constraints between the projections of the same 3D points onto the 2D images \cite{hartley2003multiple}. 
%
%
%
%
%
%
%
%
Let $p$ and $q$ denote the homogeneous coordinates of the 2D projections of the same 3D point $P$ in the reference and the current frame.
They are related by the  Longuet–Higgins equation \cite{hartley2003multiple}:
\begin{equation} 
q^{t}Ep=0
\label{eq:essential_matrix}
\end{equation}
where the matrix $E$ is known as the essential matrix, which depends on the camera rotation and translation between the two frames and the camera's intrinsic parameters. As described previously, we can use RANSAC to determine the best $E$ matrix given the set of corresponding features points in the two frames.

It is well-known   \cite{nister2004efficient} that  we can use singular value decomposition (SVD) of the essential matrix $E$ to determine the camera rotation matrix $R$ and translation vector $t$.  Specifically, we use SVD to obtain matrix $U$ and $V$ so that:
\begin{equation} 
E=UDV^{T}
\end{equation}
The rotation $R$ and translation $t$ can be computed from $U$ and $V$ as:
\begin{equation} 
R = UWV^{T}, \;\;t=U_3, \;\; {\rm with} \;\; W = 
\left [ 
  \begin{array}{ccc} 
    0 & -1 & 0\\ 
    1 & 0 & 0\\
    0 & 0 & 1\\
  \end{array}
\right ]
\end{equation}

The results are algebraically correct also with $-t$ and $W^T$, so we try all possible solutions on the matching descriptors to choose the $R$ and $t$ that leads to the least fitting error for Eq.~(\ref{eq:essential_matrix}). For implementation, we use the openCV library function \cite{pulli2012real} to calculate the essential matrix and camera rotation and translation.

Once the camera translation $t$ is determined, we update the target object location by the estimated camera translation, i.e., $o' = o-t$,
where $o$ is the object location in the reference frame and $o'$ is its location in the current frame,  relative  to the camera center and hence the user. The straight line connecting the object location $o'$ in the ground plane (i.e. the $X$ and $Z$ coordinate) and the user is the updated path.\footnote{Here we assume that there is an open space between the target and the user for simplicity. In practice, more sophisticated algorithms that detect obstacles between the target and the user and plan the path accordingly are needed. In this work, we focus on the visual processing components.}  On the other hand, the  camera translation  $t$ indicates the direction of the user's latest movement between the current frame and the last frame. We evaluate the angle between $t$ and $o'$. If the angle is larger than a pre-defined threshold, our system will sent out a friendly alert message to the user. 


\subsection{Object Localization} \label{sec3}
In this section, we introduce our object localization module, which aims to determine the 3D coordinate of the target object at the start of the navigation. Given that we only have a monocular camera, one potential option is to use a deep-learning model for determining the depth from 2D images. This is however computationally demanding. Instead, we take advantage of the fact that we have a video sequence captured while the user is moving, and use the two adjacent video frames to determine the object location.
Specifically, we first determine the 2D coordinates of the object center in the two initial frames 
 and the camera motion between the two frames. We then determine the  3D coordinate of the object center through a triangulation algorithm, as illustrated in Figure~\ref{fig3} (top).

We use the same algorithm described in Sec.~\ref{sec2} to determine the corresponding features in the first two frames and the camera motion (rotation and translation) between the two frames, except that, for  the first frame, we only perform feature extraction within the bounding  box of the detected object. We use the centroid of the 2D coordinates of all the feature points  in the object region in the first frame,  as the  coordinate of the object center in the first frame, denoted by $p$. Similarly, we determine the object center coordinate in the second frame, denoted by $q$, using feature points that correspond to the  features  belonging to the object in the first frame.

Given the camera rotation $R$ and translation $t$ and the 2D positions of the object center, $p$ and $q$, in  their homogeneous representations, we utilize triangulation \cite{hartley2003multiple} to obtain the 3D coordinate of the object center $P$ (in the homogeneous representation) with respect to the camera center in the first frame. Specifically, given the camera pose $R$ and $t$, we compute the projection matrix $J_1$ for the first frame and $J_2$ for the second frame:
\begin{equation} 
J_1 = K \cdot [I, 0],  \;\;
J_2 = K \cdot [R, t]
\end{equation}
where $K$ is the intrinsic matrix of camera and $I$ is the identity matrix. Since the cross-product between two parallel vectors equals to zero, we have:
\begin{align}
p \times (J_1 P) = 0,  \;\;
q \times (J_2 P) = 0
\end{align}
where $p = (u_1, v_1, 1)$ and $q = (u_2, v_2, 1)$. This equation can also be written as follows:
\begin{align}
\left( 
  \begin{array}{c} 
    u_1J_1^3 - J_1^1\\ 
    v_1J_1^3 - J_1^2\\
    u_2J_2^3 - J_2^1\\ 
    v_2J_2^3 - J_2^2\\
  \end{array}
\right) \cdot P = A \cdot P = 0 
\end{align}
 Then we apply SVD on $A$ to obtain $C, S,$ and $ D$ so that
\begin{equation} 
A=C S D^{T}
\end{equation}
The third column of matrix $D$ is $P$:
\begin{equation} 
P = (X, Y, Z, W) = D_3
\end{equation}
Finally, we can transform the homogeneous coordinate to the Cartesian coordinate using
\begin{equation} 
{\tilde P} = (X/W, Y/W, Z/W)
\end{equation}




\section{Experiments}
We carried out a set of experiments to evaluate the performance of our proposed system. We first use the KITTI odometry data to evaluate our system, where the video sequences are captured by a moving vehicle. We also run an experiment simulating a user walking towards a target object in an indoor environment and evaluate the performance of our algorithms. We describe these two experiments and their results separately.





\begin{figure}[h]
	\centering
	\begin{minipage}{0.8\linewidth}
		\centering
		\includegraphics[width=0.8\linewidth]{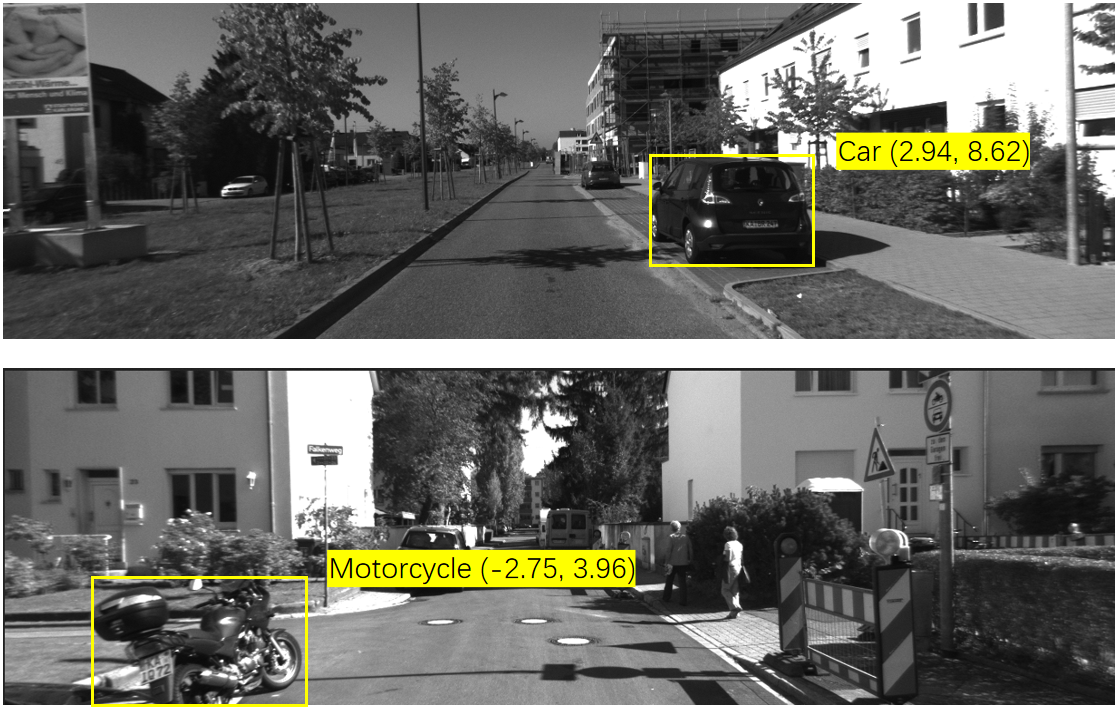}
		\label{chutian1}
	\end{minipage}
	
	\begin{minipage}{0.8\linewidth}
		\centering
		\includegraphics[width=0.8\linewidth]{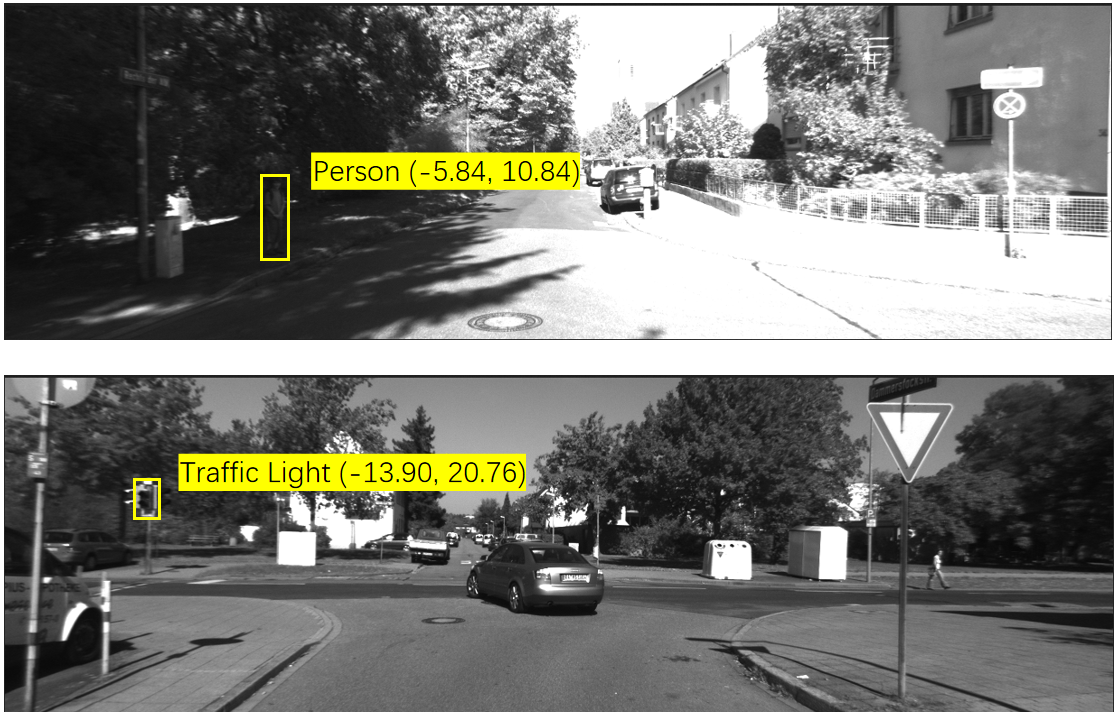}
		\label{fig_exp1}
	\end{minipage}
	\caption{Example results of object detection and object localization for the KITTI dataset. Yellow rectangles denote the bounding box of the detected objects (car and motorcycle). We also show the estimated  location  $(X, Z) $ of the detected objects. Since we are only interested in the ground position of the objects, we only show the  $X$ and $Z$ coordinate for visualization. }
		\label{fig_exp1}
\end{figure}



\subsection{Experiment with  the KITTI dataset} \label{exp2}


\subsubsection{KITTI Dataset:}

The odometry benchmark from the KITTI dataset contains 11 sequences from a car driven around a residential
area with accurate ground truth from GPS and a Velodyne
laser scanner. We choose the car, the motorbike, the pedestrian and the traffic light as possible  target objects.  We extract four video sequences from the KITTI dataset each containing a target object. Specifically, for video 1, we use  frames 3-18 in Sequence 06 of KITTI visual odometry dataset and select the car as the target object. For video 2, we use frames 2360-2370  from sequence 08 and select the motorcycle as the target object. For video 3, we use  frames 3416-3431 in Sequence 08 and select the person as the target. For video 4, we use  frames 3970-3980 in Sequence 08 and select the traffic light as the target. 

To generate the ground truth for object location, we use the corresponding 3D scan of velodyne laser data in each frame for reference. Specifically, we annotate a 3D bounding box for each object of interest  and calculate the centroid of the 3D coordinates of all points in the bounding box as the ground truth object location.

\begin{table}
\small
\parbox{.45\linewidth}{
\centering
\begin{tabular}{c|c}
    \hline                  
    Data     &Mean Absolute Error\\
    \hline
    Car& 0.23     \\
    Motorcycle& 0.27      \\
    Person& 0.14      \\
    Traffic Light & 0.67       \\
    \hline
    Mean & 0.39       \\ \hline
  \end{tabular}
\caption{ Accuracy of object localization for 4  videos on the KITTI odometry dataset. MAE in meter. }
  \label{table2}
}
\hfill
\parbox{.49\linewidth}{
\centering
\begin{tabular}{l|ll}
    \hline                  
    Data     &MAE &RMSE\\
    \hline
    Video 1 & 0.056 &  0.091    \\
    Video 2 & 0.052 & 0.085     \\
    Video 3 & 0.063 &  0.094      \\
    Video 4 & 0.055 &  0.092      \\
    \hline
    Mean & 0.056 & 0.090     \\ \hline
  \end{tabular}
\caption{Accuracy of trajectory estimation for 4 videos in the KITTI odometry dataset. MAE and RMSE in meter }
  \label{table1}
}
\end{table}

For evaluation of trajectory estimation, we use the  root mean squared error (RMSE) and mean absolute error (MAE) between the predicted translational movement and ground truth camera movement between two frames, considering only the X- and Z- coordinate. For evaluation of object localization, we use the mean absolute error (MAE) between the estimated object location and ground truth location. Lower values indicate better performance.

\subsubsection{Results:}

 We report the MAE between the predicted object location and ground truth location to validate the effectiveness of our object localization module in Table \ref{table2}.Even when estimating small objects that are far away such as the   traffic light, our system still achieves a small error of 0.67 meter. 
  
 Table \ref{table1} reports the  RMSE and MAE between the predicted trajectory between two successive frames 
 and ground truth trajectory.
 Our system is able to estimate the trajectory accurately and achieve promising results with 0.056 meter for MAE and 0.090 meter for RMSE on average over the 4 videos. 
 
  In addition to the quantitative results  discussed above, we also show example visual results in Figure \ref{fig_exp1}. Our system can successfully detect the bounding boxes and categories of the target objects by the  object detection module and estimate the object location by the object localization module.

\begin{figure}[t] 
    \centering
    \includegraphics[width=9cm]{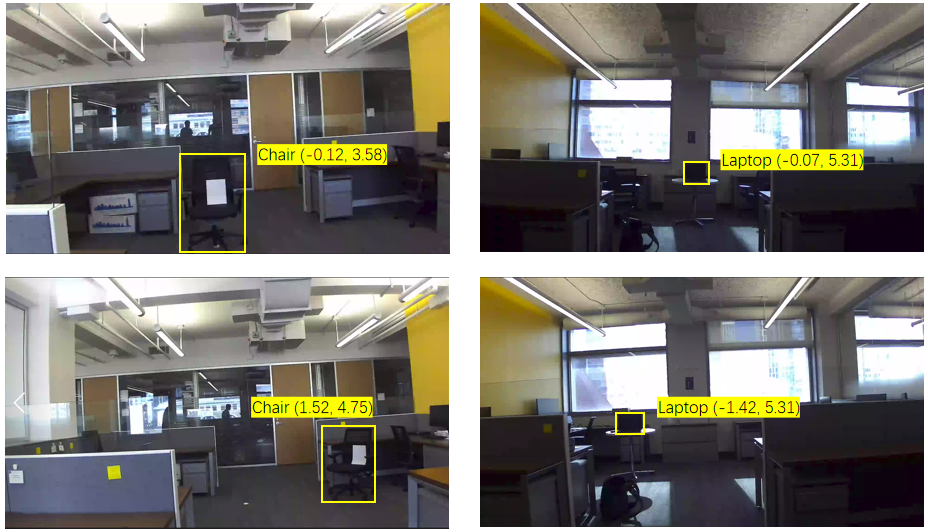}
    \caption{Examples of target object detection and object localization for our dataset. Yellow rectangles denote the bounding boxes of the detected objects (car and motorcycle). We also show the estimated  location  $(X, Z) $ of the detected objects. 
    }
    \label{fig_exp3}
\end{figure}


\subsection{Simulated Navigation Experiments} \label{exp3}
 \subsubsection{Experimental Setting:}  
 
We also conducted experiments to evaluate the proposed system in a simulated navigation experiment.  We record 4 videos by the ZED camera \cite{zed2} in an office room. Specifically, we select 2  objects of interest (laptop and chair) and record 2 videos while we walk towards each object. We use the trajectory captured by the positional tracking system of the ZED camera as the ground truth. To validate the effectiveness of our object localization module, we use the depth sensing system of ZED camera to obtain the ground truth of object position.  
 For each object, we design 2 different test scenarios. In the first scene, the object is located in front of the user. In the second scenario, the object is located to the left front or right front of the user. In each case, the video is captured while a user wearing the camera is walking straight to the front. 
 

\begin{table}
\small
\parbox{.45\linewidth}{
\centering
\begin{tabular}{c|c}
    \hline                  
    Data     &Mean Absolute Error\\
    \hline
    Chair 1 & 0.30     \\
    Chair 2 & 0.33      \\
    Laptop 1 & 0.18       \\
    Laptop 2 & 0.20      \\
    \hline
    Mean & 0.25       \\ \hline
  \end{tabular}
\caption{Accuracy of object localization using sequences in our dataset. MAE in meter  between the ground truth location and the estimated location in $X$ and $Z$.}
\label{table4}
}
\hfill
\parbox{.45\linewidth}{
\centering
\begin{tabular}{l|ll}
    \hline                  
    Data     &MAE &RMSE\\
    \hline
    Video 1 & 0.094 &  0.139    \\
    Video 2 & 0.088 &  0.138     \\
    Video 3 & 0.077 &  0.122      \\
    Video 4 & 0.083 &  0.127      \\
    \hline
    Mean & 0.086  & 0.132     \\ \hline
  \end{tabular}
\caption{Accuracy of trajectory estimation  using sequences on our  dataset.  RMSE and MAE in meter between the ground truth translation and the estimated translation.}
\label{table3}
}
\end{table}

\begin{figure}[t] 
    \centering
    \includegraphics[width=10cm]{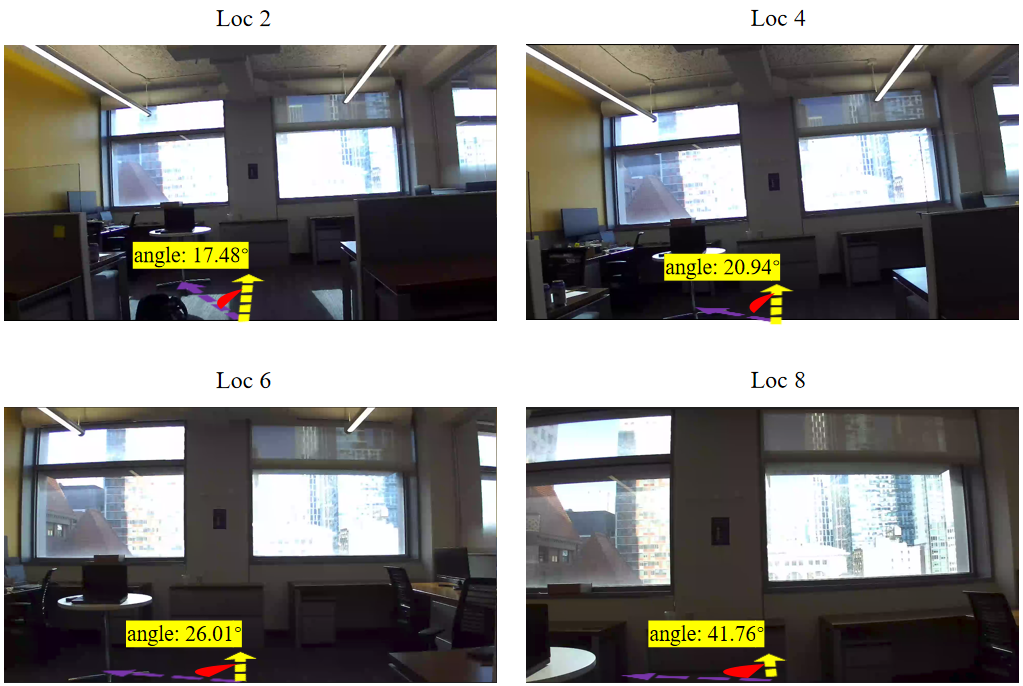}
    \caption{Examples of trajectory estimation, path update, and path correction alert on our dataset. We show  frames 2, 4, 6, and 8 in Video 4. Purple arrow denotes the updated desired path. Yellow arrow denotes the actual path of the user. As an example, in frame 8, the angle between the planned path and the user's path is larger than 30 degree, our system will sent out an alert message to the user, and suggesting the user to veer left slightly. }
    \label{fig_exp6}
\end{figure}

\begin{table*}

\caption{Running time of different modules in second. Object detection module is tested on Jetson Xavier NX with NVIDIA Volta GPU. Trajectory estimation and object localization is tested with the ARM Cortex®-A57 MPCore CPU.} \label{runtime}
\centering
\begin{tabular}{c|c|ccc}
\hline
              & \begin{tabular}[c]{@{}c@{}}Image\\ Resolution\end{tabular} & \begin{tabular}[c]{@{}c@{}}Object \\ Detection\end{tabular} & \begin{tabular}[c]{@{}c@{}}Object \\ Localization\end{tabular} & \begin{tabular}[c]{@{}c@{}}Trajectory \\ Estimation\end{tabular} \\ \hline
KITTI dataset & 1226x370                                                   & 0.18                                                        &    0.62                                                          & 0.58                                                           \\ \hline 
Our dataset   & 1920x1080                                                  & 0.18                                                        &  0.98                                                            & 0.92                                                          \\ \hline
\end{tabular}
\end{table*}

\subsubsection{Results:} 

We first examine the accuracy of   object localization module in Table \ref{table4}. Moreover, we show some sample frames with object localization results in Figure \ref{fig_exp3}. We observe that our system successfully detect the bounding box and predict the initial coordinate of the target object in the first frame.

Table \ref{table3} reports the accuracy for trajectory estimation.  We illustrate a few examples frames  in Figure \ref{fig_exp6}. As we can see from the figure, our system can provide accurate path planing and path correction. If the angle between the planned path (purple arrow) and the  user's path (yellow arrow) is larger than 30 degree,  our system will sent out an alert message to the user.


\subsection{Run time Analysis}
The run time for the three computation modules for the KITTI video and our video are summarized in Table~\ref{runtime}. We expect the run time using the Jetson  processor to be slightly higher than  using our CPU. Therefore, we expect the navigation initialization (including object detection and localization) takes less than 2 sec. and trajectory estimation takes less than 1 sec. with the Jetson processor.  This should be sufficient for real-time navigation assistance when one walks towards an object. These times can be further shortened with the optimization of the software implementation.

\section{Conclusions}
In this paper, we present a novel wearable navigation assistive system for pBLV, which augments their perceptive power so that they
can perceive objects in their surrounding environment and reach the object of interest easily. Our light-weight wearable system consists of a monocular camera mounted in  front of the chest of the user, a Jetson board for computation, and a battery.
To reduce the system cost and computation load, the system performs object detection and localization only at the start of the navigation using the first two captured frames, and then continuously update the object location relative to the user by estimating the camera motion between frames. This is akin to visual SLAM for tracking the pose of a moving camera, but here we use the visual SLAM approach to update the user's location and correspondingly the object location relative to the user.  Such continuously updated user and object locations then enable real-time navigation path update and feedback to the user. 

Our experimental results on  the KITTI odometry video dataset and simulated indoor navigation videos dataset demonstrate that the proposed system can accurately detect and localize the target object at the start of the navigation, and estimate the user movement continuously, with an error well within 0.5 meter, both outdoor and indoor.\footnote{ Note that in KITTI video and our video, the camera motion between successive frames is relatively small, leading to very small motion estimation error as well.  For the intended navigation application,  such analysis only need to be run between frames with a  larger interval, and hence larger errors are likely, but we expect them to on the same order as the localization error, which is within 0.5 meter.}
  The system is entirely vision-based and does not need other sensors for navigation (e.g. IMUs and range sensors), and the computation can be run  with the Jetson processor in the wearable system to facilitate real-time navigation assistance. Such a system holds great promise for assisting pBLV in their daily living. Future research may develop a system where the video is uploaded to an edge server for conducting all computation tasks, to further reduce the wearable system weight \cite{yuan2022network}.

\paragraph{Acknowledgments:}
Research reported in this publication was supported in part by the NSF grant 1952180 under the Smart and Connected Community program, the National Eye Institute of the National Institutes of Health under Award Number R21EY033689, and  DoD grant VR200130 under the Delivering Sensory and Semantic Visual Information via Auditory Feedback on Mobile Technology” The content is solely the responsibility of the authors and does not necessarily represent the official views of the National Institutes of Health and NSF, and DoD. Yu Hao was partially supported by NYUAD Institute (Research Enhancement Fund - RE132).

\paragraph{Conflict of Interest:}
New York University (NYU) and John-Ross Rizzo (JRR) have financial interests in related intellectual property. NYU owns a patent licensed to Tactile Navigation Tools. NYU, JRR are equity holders and advisors of said company.

\clearpage
%
%
\bibliographystyle{splncs04}
\bibliography{egbib}
\end{document}